\documentclass[10pt,twocolumn,letterpaper]{article}

\usepackage{iccv}
\usepackage{times}
\usepackage{epsfig}
\usepackage{graphicx}
\usepackage{amsmath}
\usepackage{amssymb}
\usepackage{booktabs}
\usepackage{multirow}
\usepackage{bbm}
\usepackage{color}
\usepackage{amsmath}
\usepackage{mathtools}
\usepackage{subcaption}

\usepackage[breaklinks=true,bookmarks=false]{hyperref}

\iccvfinalcopy 

\def\iccvPaperID{****} 

\ificcvfinal\fi

\begin{document}

\title{DF$^{2}$AM: Dual-level Feature Fusion
and Affinity Modeling for RGB-Infrared Cross-modality Person Re-identification}

\author{Junhui Yin, Zhanyu Ma$^{*}$, Jiyang Xie, Shibo Nie, Kongming Liang, and Jun Guo 
}

\maketitle
\ificcvfinal\thispagestyle{empty}\fi

\begin{abstract}
RGB-infrared person re-identification is a
challenging task due to the intra-class variations and cross-modality discrepancy. Existing works mainly focus on learning modality-shared global representations by aligning image styles
or feature distributions across modalities, while local feature from body part and relationships between person images
are largely neglected. In this paper, we propose a Dual-level (i.e., local and global) Feature Fusion (DF$^{2}$) module by learning attention for discriminative feature from local to global manner. In particular, the attention for a local feature is determined locally, i.e., applying a learned
transformation function on itself. Meanwhile, to further mining the relationships between global features from person images, we
propose an Affinities Modeling (AM) module to obtain the optimal intra- and inter-modality image matching. Specifically, AM employes intra-class compactness and inter-class
separability in the sample similarities as supervised information to model the affinities
between intra- and inter-modality samples. Experimental results show that
our proposed method outperforms state-of-the-arts by large margins
 on two widely used cross-modality re-ID datasets
SYSU-MM01 and RegDB, respectively.
\end{abstract}

\section{Introduction}

Person re-identification (re-ID)
is a
cross-camera image retrieval task, which aims to match persons of a given query
from an image gallery collected from disjoint cameras.
Many studies resort to deep metric
learning~\cite{hermans2017defense,zheng2017discriminatively}, or use classification losses
as the proxy targets to extract discriminative features~\cite{li2017person,sun2018beyond,tang2019cityflow,wu2019progressive}.
With recent progress in the generative adversarial networks
(GANs), another possibility is explore GANs as a style transformer to augment training data and improve the discriminative
capacity of model~\cite{zheng2017unlabeled,ge2018fd,liu2018pose,qian2018pose,zheng2019joint}.

Existing re-ID methods mainly treat visible images as single-modality
and degrade dramatically in real complex scenarios where person images are captured from both dark and bright lighting environments.
However, visible light cameras can not work at night.
Fortunately, some surveillance devices like infrared cameras can
capture the appearance characteristics of a person under poor illumination
conditions to overcome these difficulties.
This yields popular research interest on RGB-IR cross-modality matching, which is more
challenge due to the large discrepancy between two modalities compared with the RGB single modality. For instance, RGB images contain some discriminative cues like colors while these information are missing in infrared images.

\begin{figure}
\begin{center}
\includegraphics[width=1\columnwidth]{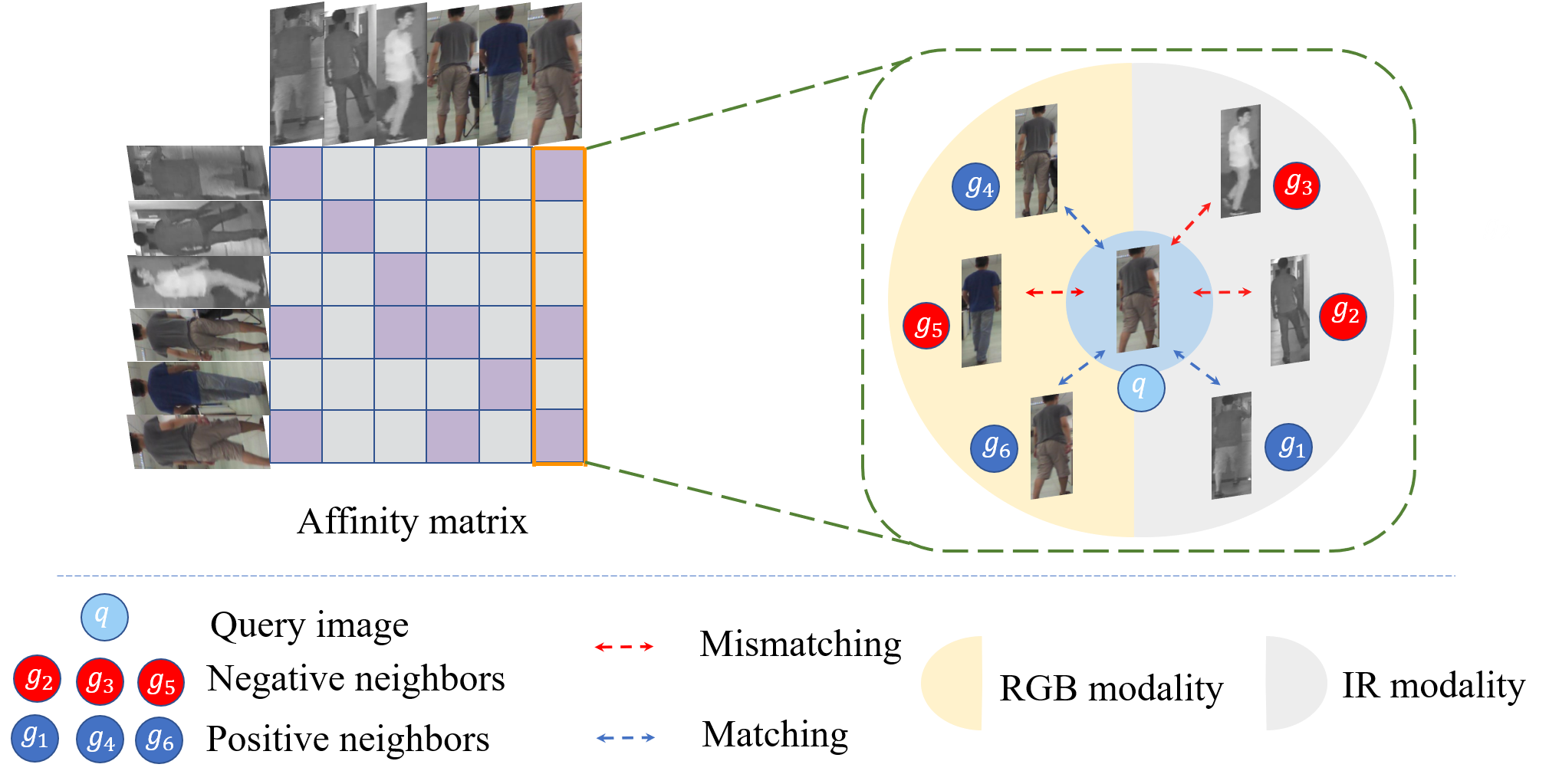}
\end{center}
   \caption{\small Affinity modeling (AM) infers cross-modality
sample similarities by exploiting intra-class compactness and inter-class separability in the sample similarities.
 In training batch, each training image can be considered as query image and it treats all the training samples as its neighbours, where each image accepts the structure
information from all the neighbours.
}
\label{fig:intro}
\end{figure}

Recently, many studies resort to two typical approaches to address
the aforementioned challenges in cross-modality re-ID.
The first approaches~\cite{wu2017rgb,ye2018hierarchical,ye2018visible} attempt to reduce the cross-modality discrepancy with feature-level constraints
like aligning feature distribution of images.
The other approaches~\cite{wang2019rgb,wang2019learning,wang2020cross} are at input-level using GANs to transfer images from one modality to another while preserving the identity information as much as possible.
The two approaches mainly focus on reducing discrepancy across modalities, whereas there is still the challenge of appearance variations in a single RGB or IR modality, including background
clutter, viewpoint variations, occlusion, etc.

To address the above problem, we propose to learn attention for
discriminative feature from local to global manner.
The critical idea behind it lies in different parts of a person containing different
discriminative information.
The network model can still capture useful information from the upper body using attention mechanism, regardless of a pedestrian's lower body  occluded by something (\emph{e.g}, a bicycle).
Specifically,
we propose a $local$ $attention$: the attention for a local feature is determined locally, \emph{i.e.}, applying a learned transformation function on itself, where the refined part-aggregated features consider the importance between different body parts. However, such local strategies do not fully exploit the feature information from a global view.
Our solution is to use the global feature information from feature maps using global average pooling (GAP), which is named $global$
 $attention$.
In this way, we consider both the global feature and its part information
to determine importance between different body parts of a person from global and local
views. This is also consistent with the perception
of human in finding discriminative cues: making a
comparison and then determine the importance.

The aforementioned method processes each sample independently, ignoring the relationships between person images.
Thus, we present a novel and efficient similarity inference
to obtain the optimal intra- and inter-modality image matching.
It utilizes intra-class compactness and inter-class separability in the sample similarities as supervised information to model the affinities between intra- and
inter-modality samples.
In particular, every sample contains some structural information and propagates the information to its neighbors using the pairwise relations, which is shown in Figure~\ref{fig:intro}. This neighbor reasoning scheme
can compensate for the lack of specific information existing in the same person's different images and further
enhance the robustness of the learned feature from object-level.

In our proposed method, contextual information for RGB-IR cross-modality re-ID
consists of dual levels. The lowest is the patch-level where the appearance variation (\emph{e.g.}, data occlusion problem) is mitigated with a
weighted sum over important body parts to assist more accurate information of object (\emph{i.e.}, person).
At the object level, coexistence
of objects provides strong hints on identification of the same person.
Experimental results show that our proposed method can surpass state-of-the-arts by large margins
on two widely used cross-modality
re-ID datasets
SYSU-MM01~\cite{wu2017rgb} and RegDB~\cite{nguyen2017person}, respectively.

In summary,
the contributions of our work include:

$\bullet$ We propose to learn the attention for discriminative representation by taking both local and global views of features.

$\bullet$ We design an efficient mutual neighbor reasoning to capture long-range dependency of objects,
by modeling affinity between intra- and inter-modality images.

$\bullet$ Our proposed method achieves significant performance improvement
over the state-of-the-arts on
on the two most popular benchmark datasets.

\begin{figure*}
\begin{center}
\includegraphics[width=2\columnwidth]{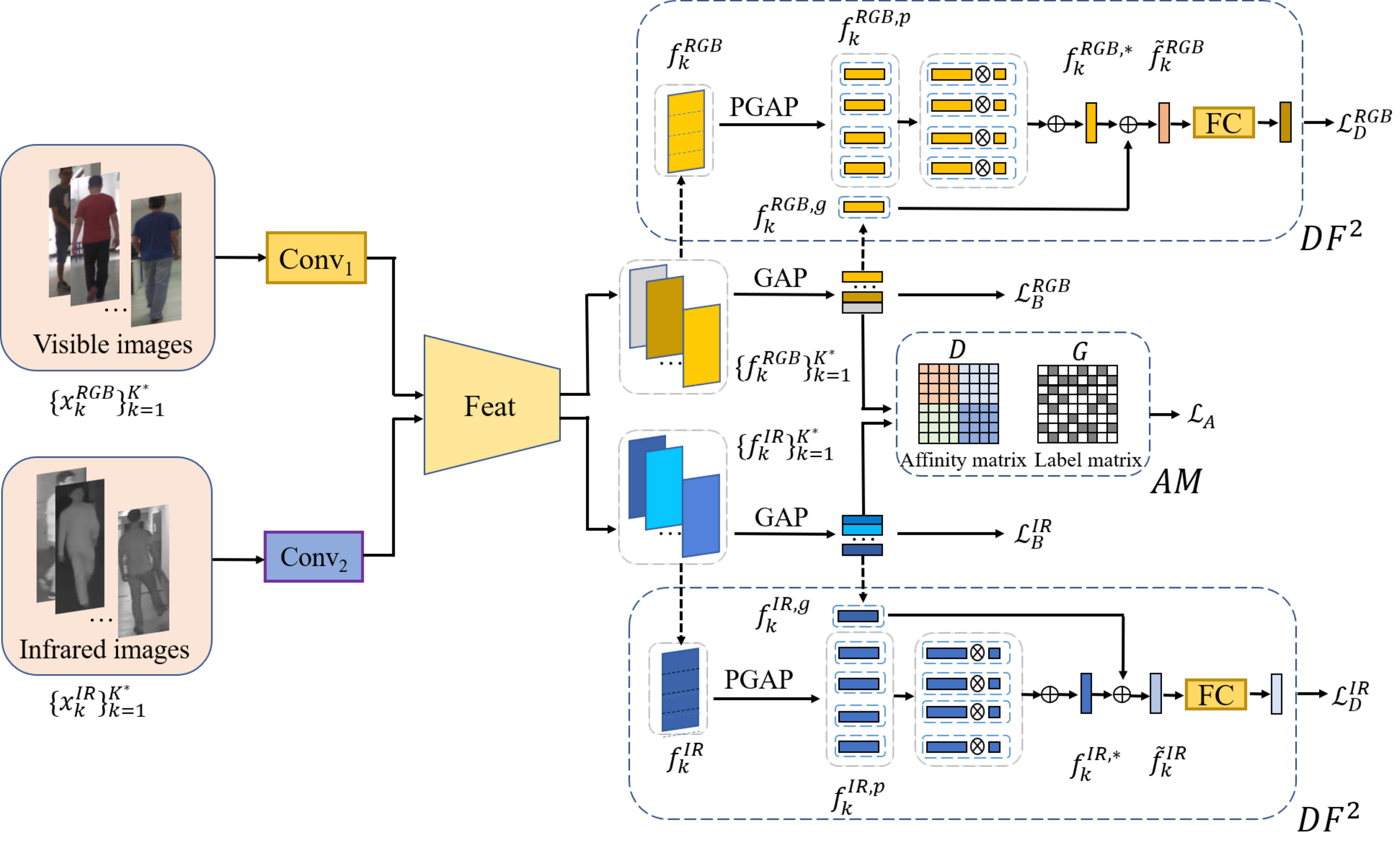}
\end{center}
   \caption{\small The architecture of our DF$^{2}$AM method. The entire framework includes two important components: the weighted-part and global feature fusion and affinities modeling for intra-
and inter-modality feature matching. Our goal is to learn discriminative
features and enhance the robustness of the learned features from patch-level to object-level.
}
\label{fig:overview}
\end{figure*}

\section{Related Work}

\textbf{RGB-RGB single-modality person re-Identification}.
Conventional person re-ID research is a RGB-RGB single modality re-ID to address the problem of matching pedestrian across non-overlapping cameras. The challenge of re-ID lies in how to learn discriminative features from person images where there are the large intra-class and small inter-class variations caused by diversity of poses, illumination conditions, viewpoint occlusion, etc. To address the aforementioned challenges, many deep re-ID methods~\cite{guo2019beyond,sun2019perceive,sun2018beyond,liu2015spatio} have been proposed. Some of them resort to a partial feature learning~\cite{guo2019beyond,sun2019perceive} and focus on the powerful network structures to align the body parts~\cite{sun2018beyond,liu2015spatio}. Other methods try to discard the appearance variant out in metric space using loss functions, which contains  contrastive loss~\cite{hermans2017defense}, triplet loss~\cite{hermans2017defense}, quadruplet loss~\cite{chen2017beyond}.  Recent graph-based methods~\cite{shen2018person,wu2019unsupervised} consider connections between sample pairs. However, these methods are developed for single-modality re-ID but not for the cross-modality re-ID due to the large discrepancy across modalities.

\textbf{RGB-IR cross-modality person re-Identification}. The large discrepancy of cross-modality re-ID comes not only from appearance variations but also from cross-modality variation between RGB and IR images. Existing studies for cross-modality re-ID can be mainly summarized into two categories of methods. The first category~\cite{wu2017rgb,ye2018hierarchical,ye2018visible} attempts to align the feature distribution of training images in representation space. The work~\cite{wu2017rgb} focus on how to design one-stream networks such as a deep zero-padding network for evolving domain-specific nodes. The two-stream network with modality-specific~\cite{ye2018hierarchical} and top-ranking loss~\cite{ye2018visible} are developed to learn multi-modality
representations. In~\cite{dai2018cross}, a generative adversarial training method is proposed to jointly discriminate the identity and modality. ~\cite{hao2019hsme} design a hyper-sphere manifold embedding model to learn discriminative representations from different modalities. The second approach instead uses cross-modality generative adversarial network (GAN) to transfer person images style from one modality to another.~\cite{kniaz2018thermalgan} collects a new ThermalWorld dataset and propose a ThermalGAN framework for color-to-thermal image translation.~\cite{wang2019learning} further considers dual-level discrepancy and use a bi-directional cycle GAN~\cite{zhu2017unpaired} to generate unlabeled images as data augmentation. A hierarchical disentanglement method~\cite{choi2020hi} is proposed to disentangle ID-discriminative factors and ID-excluded factors simultaneously by using pose- and illumination-invariant features from cross-modality images.

\textbf{Attention mechanisms}.  Humans does not attempt to process a whole scene of the data at once. Instead, they selectively use an salient part information to make a decision~\cite{itti1998model,mnih2014recurrent}. The above process is called attention mechanism and is actively used in many tasks including image captioning~\cite{xu2015show,chen2017sca}, transfer learning~\cite{zagoruyko2016paying}, object localization~\cite{zhang2018self}. Furthermore, the self-attention mechanism~\cite{vaswani2017attention} is proposed to draw global dependencies of inputs. Recently, various methods~\cite{wang2017residual,wang2018non,hu2018squeeze,woo2018cbam} use the self-attention mechanism to improve the performance of the classification model. For person re-ID, attention is used to capture spatial and temporal characteristics of pedestrian sequences from different video frames~\cite{xia2019second,li2018diversity,li2018diversity}. However, the application of these methods is limited for the cross-modality re-ID due to the different camera environments and large visual appearances change. In this work, we use attention mechanism to focus on important local features instead of processing all the data equally for cross-modality re-ID.

\section{Methodology}

In this section, we describe details of the proposed DF$^{2}$AM approach for cross-modality re-ID. We first revisit the baseline single-modality model and introduce a more efficient way for intra- and inter-modality feature matching. Then, the details of the proposed DF$^{2}$AM, founded on the above finding, are presented for learning discriminative features, and enhancing the robustness of the learned feature from patch-level to object-level.


\subsection{Single-modality Person Re-identification Revisit}

We first present the baseline single-modality re-ID model, which offers a promising way to learn discriminative global features. The training process can be treated as a conventional classification problem~\cite{zheng2016person}. To learn the feature embedding, the baseline usually learn the parameters with manual annotations where extracted feature $f_{k},k=1,\cdots,K$ is associated with a one-hot label $y_{k}$. The classification procedure is achieved by minimizing a cross-entropy loss $\mathcal{L}_{ID}$. Meanwhile, hard-mining triplet loss $\mathcal{L}_{BH}$~\cite{hermans2017defense} is used to optimizes the triplet-wise relationships among different person images. Thus, the baseline re-ID model is optimized by minimizing the following loss function as
\begin{equation}
\mathcal{L}_{B}=\mathcal{L}_{ID}+\mathcal{L}_{BH},
\end{equation}
where
\begin{footnotesize}
\begin{align}
L_{ID}&=-\frac{1}{K}\sum_{i=1}^{K}\log p(y_{k}|f_{k}),\\
L_{BH}&=
\overbrace{\sum_{i=1}^{N}\sum_{a=1}^{M}}^{\text{all anchors}}\Big[m+\overbrace{\max_{p=1,\cdots,M}d(f_{a},f_{p})}^{\text{hardest positive}}
-\underbrace{\max_{\substack{j=1,\cdots,N \\ n=1,\cdots,M \\ j\neq i}} d(f_{a},f_{n})}_{\text{hardest negative}}\Big]_{+},
\end{align}
\end{footnotesize}
Here, $d$ is a metric function (\emph{i.e.}, European distance) measuring distances in the embedding
space,
$K=MN$ is the number of images in single modality, $N$ is randomly selected identity number, and $M$ is randomly sampled image number of each identity.
$p(y_{i}|f_{k})$ is the predicted probability that the encoded feature $f_k$ belongs to its identity $y_k$, which is obtained by a classifier.


\subsection{The overall framework}

To explore richer
visual patterns for cross-modality re-ID, we propose DF$^{2}$AM method and integrate it with a conventional re-ID network. The DF$^{2}$AM is fulfilled from the perspective of patch-level and object-level, and is deployed as dual-level feature fusion and affinity modeling modules.
The learning procedure is fulfilled by optimizing a joint objective function, as

\begin{equation}\label{overall loss}
    \mathcal{L}_{Final}=\mathcal{L}_B+\lambda\mathcal{L}_D+\zeta\mathcal{L}_A,
\end{equation}
where $\mathcal{L}_B=\mathcal{L}_B^{RGB}+\mathcal{L}_B^{IR}$. Here, $\mathcal{L}_B^{RGB}$ and $\mathcal{L}_B^{IR}$ are the baseline loss for
RGB and IR modality respectively, $\mathcal{L}_D$ denotes the classification loss for dual-level feature fusion module,
and $\mathcal{L}_A$ is the affinity modeling loss. $\lambda$ and $\zeta$ are the regularization
factors.

As shown in Figure~\ref{fig:overview},
we first feed visible image batch $X^{\text{RGB}}=\{x_k^{\text{RGB}}\}_{k=1}^{K^{*}}$ and infrared image batch $X^{\text{IR}}=\{x_k^{\text{IR}}\}_{k=1}^{K^{*}}$ into different convolutional layers to capture modality-specific low-level feature patterns, where $2K^{*}$ is the batch size. Then,
we use the shared feature extractor
 (\emph{i.e.}, convolutional layers) $Feat$ to transform the specific features
onto a common representation space to acquire modality-sharable high-level features,
formulated as
$$
f_k^{\text{RGB}}=Feat(Conv_{1}(x_k^{RGB})),$$
$$f_k^{\text{IR}}=Feat(Conv_{2}(x_k^{IR})).$$
Here, $f_k^{\text{RGB}}\subseteq \mathbb{R}^{C\times H\times W}$ and $f_k^{\text{IR}}\subseteq \mathbb{R}^{C\times H\times W}$ for visible and infrared images, respectively.
Note that $C$ is the number of channels,
$H$ and $W$ are height and width, respectively.
With the obtained high-level features, intra- and inter-modality feature learning are required to be undertaken. For the intra-modality feature learning, dual-level feature fusion (DF$^{2}$) module learns part-aggregated feature embeddings, and combine them with global features to enhance the representative capacity of features
for intra-modality re-ID. Meanwhile, the shared feature extractor is to learn aligned features
for bridging RGB and IR modalities. For inter-modality feature matching, a similarity inference is presented to model the affinities between both intra- and inter-modality global features. This neighbor
reasoning scheme utilizes intra-class compactness and inter-class
separability in the sample similarities to enhance the robustness of the learned feature
from object-level.

%

%

\subsection{Dual-level Feature Fusion}\label{ssec:localattention}

Previous cross-modality re-ID models~\cite{hao2019hsme,wang2020cross} commonly focus on constructing feature- or image-level constraints for reducing distribution discrepancy in same identities. However, the challenge of appearance variations, including background clutter, viewpoint variations, and occlusion, are not overcome by only using the global features. In this case, we propose the local attention mechanism, which refines part-aggregated features by a learned transformation function on itself, to consider the importance between different body parts of a person.

We take the feature $f_k^{\text{RGB}}$ from the RGB modality as an example. In our local attention mechanism, a patch-wise average pooling (PAP) is used to extract $P$ local features $f_k^{\text{RGB},p}\subseteq \mathbb{R}^{C},p=1,\cdots,P$ (assuming $P$ is a factor of $H$) by
\begin{equation}
    f_k^{\text{RGB},p}=\frac{P}{WH}\sum_{w=1}^W\sum_{h=\frac{(p-1)H}{P}+1}^{\frac{pH}{P}}f_{k,:,h,w}^{\text{RGB}},
\end{equation}
where PAP first splits the feature maps $f_k^{\text{RGB}}$ into $P$ horizontal feature spatial parts and then generates local features
$f_k^{\text{RGB},p}$ by compressing spatial parts using global average pooling (GAP).

To obtain a discriminative part-aggregated feature $\tilde{f}_k^{\text{RGB}}$, we compute a weighted summation of local features from body parts together
with learnable local attention as weight $\omega=(\omega_1,\cdots,\omega_P)^{\text{T}}$. In summary, it is formulated by
\begin{equation}\label{eq:localattention}
    f_k^{\text{RGB},*}=\sum_{p=1}^P\tilde{\omega}_p f_k^{\text{RGB},p},
\end{equation}
where $\tilde{\omega}_p=\frac{e^{\omega_p}}{\sum_{p'=1}^{P}e^{\omega_{p'}}}$.

Although the aforementioned local attention mechanism assigns weights for local features, such local strategy cannot fully exploit the feature information from a global view and affect the representative capacity of feature, which is discussed in the experiments. Thus, we combine global feature with local feature as
\begin{equation}
    \tilde{f}_k^{\text{RGB}}=\text{BN}(f_k^{\text{RGB,g}})+f_k^{\text{RGB},*},
\end{equation}
where $\text{BN}(\cdot)$ is batch normalization and $f_k^{\text{RGB,g}}$ represents the GAP output of the input
feature map $f_k^{\text{RGB}}$. For the infrared modality, we can also obtain the discriminative representation $\tilde{f}_k^{\text{IR}}$ by using $f_k^{\text{IR}}$ in the same way. Finally, the loss for $DF^{2}$ can be expressed by
\begin{equation}
\small
\begin{aligned}
    \mathcal{L}_{D}&=\mathcal{L}_{D}^{RGB}+\mathcal{L}_{D}^{IR}  \\
    =&-\frac{1}{K^{RGB}}\sum_{k=1}^{K^{RGB}}\log p(y_{k}|\tilde{f}_k^{\text{RGB}}) \\
    &-\frac{1}{K^{IR}}\sum_{k=1}^{K^{IR}}\log p(y_{k}|\tilde{f}_k^{\text{IR}}),
\end{aligned}
\end{equation}
where
$K^{RGB}$ ($K^{IR}$) is the
number of images in visible (infrared) modality, and each of features $\tilde{f}_k^{\text{RGB}}$ and $\tilde{f}_k^{\text{IR}}$ corresponds to an
identity label $y_{k}\in\{1,\cdots,N\}$.

\subsection{Affinity Modeling}\label{ssec:similarityinference}

A common strategy to align feature distribution of images is utilizing hard-mining triplet loss~\cite{hermans2017defense}.
For each chosen anchor, it is required for selecting the hard positive/negative exemplars from within a mini-batch. The above strategy is complex and required for additional computing resources.
To better align feature distribution of images across intra- and inter-modality,
 we propose a simple and efficient similarity inference to obtain the optimal intra- and inter-modality image
matching. It utilizes intra-class compactness and inter-class
separability in the sample similarities as supervised information to model the affinities
between intra-and inter-modality samples.
Here we aim to ensure that an image of a specific person is closer to all positive images of the same person than any (negative) image of any other person. In addition, our method is simple and efficient since no hard pairs are mined and all the pairs are utilized in training.

\textbf{Affinity matrix construction}. We first use the encoded global features $f_k^{\text{RGB,g}}$ and $f_k^{\text{IR,g}}$ from two modalities to model the pair-wise affinities (affinity matrix $D$), which is defined as
\begin{equation}\label{1}
    D=\begin{pmatrix}
        D_{\text{RGB,RGB}} & D_{\text{RGB,IR}} \\
        D_{\text{IR,RGB}} & D_{\text{IR,IR}}
      \end{pmatrix},
\end{equation}
where $D_{\text{RGB,RGB}}$ and $D_{\text{IR,IR}}$ are intra-modality affinity matrices of the visible and infrared modalities, respectively, and $D_{\text{RGB,IR}}$ and $D_{\text{IR,RGB}}$ are inter-modality affinity matrices.
At each training step, an identity-balanced sampling strategy are adopted for training~\cite{eccv20ddag}.
For each of $N^{*}$ different randomly selected identities, $M^{*}$ visible and $M^{*}$ infrared images are randomly sampled, resulting in
batch size $2K^{*}$ is equal to $2N^{*}M^{*}$ and $D_{a,b},a,b\in\{\text{RGB},\text{IR}\}$ are $K^{*} \times K^{*}$-dimensional.
The elements of the sub-matrices $D_{a,b},a,b\in\{\text{RGB},\text{IR}\}$ are calculated by
\begin{equation}\label{2}
D_{a,b}^{ij}=\left|\left|\frac{f_i^{a\text{,g}}}{||f_i^{a\text{,g}}||_2}-\frac{f_j^{b\text{,g}}}{||f_j^{b\text{,g}}||_2}\right|\right|_2 \in [0,
+\infty),
\end{equation}
where $||\cdot||_2$ is $L_2$-norm. It is noted that other distance metrics, such as cosine similarity, can be also used. In this work, we adopt the Euclidean distance between normalized global features and smaller distance value means more similar.

\textbf{Ground truth affinity matrix}. According to label information of each image, we construct a ground truth affinity, which corresponds to the affinity matrix and performs as the ground truth labels of the similarity inference. The label matrix is defined as a binary matrix $G=[G^{ij}],i,j=1,\cdots,2K^{*}$, where $G^{ij}=1$ if the $i^{th}$ and the $j^{th}$ samples belong to one same identity, otherwise zero.


Thus, our aim is that the affinity elements of negative pairs are as close to one as possible and that of positive pairs are as close to zero as possible.
This can be achieved by minimizing the following mean $L_{1}$ error between affinity matrix and its ground truth,
\begin{equation}
    \mathcal{L}_1=||D-(\mathbbm{1}-G)\cdot \delta||_1\rightarrow0,
\end{equation}
where $\mathbbm{1}$ is a matrix whose elements are all $1$ and $\delta$ is a suitable large value.

Theoretically, the above consistency condition is a necessary but not sufficient requirement for discriminative embeddings. This may in practice induce model converge to bad local minima early in training, which is also proved by our experiment. The main reason is
all the positive and negative pairs are still treated equally when some pairs' distances are suitable enough.
To address this, our solution is to relax the learning objective. We expect the affinity elements of negative pairs are larger than that of positive pairs, which can be achieved by constraining a suitable margin $m$ between the distances of positive and negative pairs.
Thus, we modify the objective loss function to
\begin{equation}
    \mathcal{L}_A=\sum_{i=1}^{2K^{*}}\sum_{j=1}^{2K^{*}}[D^{ij}\otimes G^{ij}-(D^{ij}-m)\otimes(1-G^{ij})]_{+},
\end{equation}
where $m$ is a margin and $\otimes$ presents Hadamard product.

This loss makes sure that the affinity elements of negative pairs are large at leat a
margin $m$. The advantage of this formulation is that, compared with hard-mining triplet loss~\cite{hermans2017defense}, we can easily optimize this loss over the whole dataset without a long enough training and all points of the same class merely need to be closer to each other than to any point
from a different class.

%

\begin{table*}[!t]
  \centering
    \begin{tabular}{lcccccccccc}
    \toprule
    \multicolumn{1}{c}{Setting} &       & \multicolumn{4}{c}{All Search} &       & \multicolumn{4}{c}{Indoor Search} \\
\cmidrule{1-1}\cmidrule{3-6}\cmidrule{8-11}   \multicolumn{1}{c}{Method} &       & \multicolumn{1}{c}{r = 1} & \multicolumn{1}{c}{r = 10} & \multicolumn{1}{c}{r = 20} & \multicolumn{1}{c}{mAP} &       & \multicolumn{1}{c}{r = 1} & \multicolumn{1}{c}{r = 10} & \multicolumn{1}{c}{r = 20} & \multicolumn{1}{c}{mAP} \\
    \midrule
    One-stream~\cite{wu2017rgb}(ICCV'17) &       & 12.04 & 49.68 & 66.74 & 13.67 &       & 16.94 & 63.55 & 82.10  & 22.95 \\
    Two-stream~\cite{wu2017rgb}(ICCV'17) &       & 11.65 & 47.99 & 65.50  & 12.85 &       & 15.60  & 61.18 & 81.02 & 21.49 \\
    Zero-Pad~\cite{wu2017rgb}(ICCV'17) &       & 14.80  & 54.12 & 71.33 & 15.95 &       & 20.58 & 68.38 & 85.79 & 26.92 \\
    TONE~\cite{ye2018hierarchical}(AAAI'18)  &       & 12.52 & 50.72 & 68.60  & 14.42 &       & 20.82 & 68.36 & 84.46 & 26.38 \\
    HCML~\cite{ye2018hierarchical}(AAAI'18)  &       & 14.32 & 53.16 & 69.17 & 16.16 &       & 24.52 & 73.25 & 86.73 & 30.08 \\
    cmGAN~\cite{dai2018cross}(IJCAI'18) &       & 26.97 & 67.51 & 80.56 & 31.49 &       & 31.63 & 77.23 & 89.18 & 42.19 \\
    BDTR~\cite{ye2019bi}(TIFS'19)  &       & 27.32 & 66.96 & 81.07 & 27.32 &       & 31.92 & 77.18 & 89.28 & 41.86 \\
    eBDTR~\cite{hao2019hsme}(AAAI'19) &       & 27.82 & 67.34 & 81.34 & 28.42 &       & 32.46 & 77.42 & 89.62 & 42.46 \\
    HSME~\cite{hao2019hsme}(AAAI'19)  &       & 20.68 & 32.74 & 77.95 & 23.12 &       & - & - & - & - \\
    D$^{2}$RL~\cite{wang2019learning}(CVPR'19)  &       & 28.90  & 70.60  & 82.40  & 29.20  &       & - & - & - & - \\
    MAC~\cite{ye2020cross}(TIP'20)   &       & 33.26 & 79.04 & 90.09 & 36.22 &       & 36.43 & 62.36 & 71.63 & 37.03 \\
    MSR~\cite{feng2019learning}(TIP'19)   &       & 37.35 & 83.40  & 93.34 & 38.11 &       & 39.64 & 89.29 & 97.66 & 50.88 \\
    AlignGAN~\cite{wang2019rgb}(ICCV'19) &       & 42.40  & 85.00    & 93.70  & 40.70  &       & 45.90  & 87.60  & 94.40  & 54.30 \\
    Xmodal~\cite{li2020infrared}(AAAI'20) &       & 49.92 & 89.79 & 95.96 & 50.73 &       & - & - & - & - \\
    DDAG~\cite{eccv20ddag}(ECCV'20)  &       & 54.75 & 90.39 & 95.81 & 53.02 &       & 61.20  & 94.06 & 98.41 & 67.98 \\
    \midrule
    Ours  &       & \textbf{56.93} & \textbf{90.80} & \textbf{96.11} & \textbf{55.10} &       & \textbf{66.39}  & \textbf{94.93} & \textbf{98.55} & \textbf{71.52} \\
    \bottomrule
    \end{tabular}%
      \caption{Experimental results of the proposed DF$^{2}$AM and state-of-the-art methods on on SYSU-MM01 dataset under two different settings. Rank at r accuracy (\%) and mAP (\%) are reported.}
  \label{tab:SYSU-MM01}%

\end{table*}%

\section{Experimental Results}

\subsection{Datasets and Settings}

\textbf{Datasets}. The proposed method is evaluated over two widely used cross-modality datasets,
SYSU-MM01~\cite{wu2017rgb} and RegDB~\cite{nguyen2017person}.
SYSU-MM01 is a popular RGB-IR re-ID dataset, which contains images of 419 identities
captured in both indoor and outdoor environments.
The training set includes 22,258
RGB images and 11,909 IR images of 395 persons, and the testing set contains 96
identities, with 3,803 IR images for the query and 301 RGB images for the gallery set.
There are two different testing settings for RGB-IR re-ID: indoor-search
and all-search~\cite{wu2017rgb}. All-search mode treats images from all RGB cameras as the gallery set, while the images of
gallery set are captured by two indoor RGB cameras in the indoor-search mode.
RegDB contains 412 persons and 8,240 images, where each person has 10 RGB images and 10 IR images.
There are 4,120 images of 206 persons for training and the remaining images of 206 persons for testing.
The testing set has two evaluation modes, Visible to Infrared and Infrared to Visible, where the former is to search RGB images from a infrared image and the latter is to search IR images from a Visible image.
The stable result is obtained over 500 repetitions on a random split.

\textbf{Evaluation protocols}.
We use two standard evaluation protocols as evaluation metrics:
Cumulative Matching Characteristic (CMC) and mean average precision (mAP).
The Rank-$k$ identification rate in the CMC curve represents the cumulative rate of true
matches in the top-k position, while mAP treats person
re-identification as a retrieval task.

\textbf{Implementation details}.
We adopt ResNet-50~\cite{he2016deep} with
the last classification layer removed as the backbone network and initialize it by using parameters pre-trained on ImageNet~\cite{deng2009imagenet}.
We randomly sample $N^{*}$ identities, and then
randomly sample $M^{*}$ visible and $M^{*}$ infrared images to constitute a training batch, so that the mini-batch size is $2K^{*}=2N^{*}\times M^{*}$.
In this paper, $N^{*}$ and $M^{*}$ are
set to 8 and 4, respectively.
We resize input images to 256$\times$128 and then employ random cropping and flipping as data augmentation for them.
During training, we also use the SGD optimizer with the momentum parameter 0.9 and the initial learning rate 0.1 for 80 epoches. The learning rate decays by 0.1 and 0.01 at the 30th and 50th epoch, respectively.

\begin{table*}[!t]
  \centering
    \begin{tabular}{lrccccccccc}
    \toprule
    Setting &       & \multicolumn{4}{c}{Visible to Infrared} &       & \multicolumn{4}{c}{Infrared to Visible } \\
\cmidrule{1-1}\cmidrule{3-6}\cmidrule{8-11}    Method &       & r = 1   & r = 10  & r = 20  & mAP   &       & r = 1   & r = 10  & r = 20  & mAP \\
    \midrule
    HCML~\cite{ye2018hierarchical}(AAAI'18)  &       & 24.44 & 47.53 & 56.78 & 20.08 &       & 21.70  & 45.02 & 55.58 & 22.24 \\
    Zero-Pad~\cite{wu2017rgb}(ICCV'17) &       & 17.75 & 34.21 & 44.35 & 18.90  &       & 16.63 & 34.68 & 44.25 & 17.82 \\
    BDTR~\cite{ye2019bi}(TIFS'19)  &       & 33.56 & 58.61 & 67.43 & 32.76 &       & 32.92 & 58.46 & 68.43 & 31.96 \\
    eBDTR~\cite{ye2019bi}(AAAI'19) &       & 34.62 & 58.96 & 68.72 & 33.46 &       & 34.21 & 58.74 & 68.64 & 32.49 \\
    HSME~\cite{hao2019hsme}(AAAI'19)  &       & 50.85 & 73.36 & 81.66 & 47.00    &       & 50.15 & 72.40  & 81.07 & 46.16 \\
    D$^{2}$RL~\cite{wang2019learning}(CVPR'19)  &       & 43.4  & 66.1  & 76.3  & 44.1  &       &   -    &   -   &  -   & -  \\
    MAC~\cite{ye2020cross}(TIP'20)   &       & 36.43 & 62.36 & 71.63 & 37.03 &       & 36.20  & 61.68 & 70.99 & 36.63 \\
    MSR~\cite{feng2019learning}(TIP'19)   &       & 48.43 & 70.32 & 79.95 & 48.67 &       &   -  &   -  &   -  & -   \\
    AlignGAN~\cite{wang2019rgb}(ICCV'19) &       & 57.90  &  -  &  -   & 53.60  &       & 56.30  &   -  &   -  & 53.40 \\
    Xmodal~\cite{li2020infrared}(AAAI'20) &       & 62.21 & 83.13 & 91.72 & 60.18 &       & -   &  -   &   -  & - \\
    DDAG~\cite{eccv20ddag}(ECCV'20)  &       & 69.34 & 86.19 & 91.49 & 63.46 &       & 68.06 & 85.15 & 90.31 & 61.80 \\
    \midrule
    Ours  &       & \textbf{73.06} & \textbf{87.96} & \textbf{91.51} & \textbf{67.81} &       & \textbf{70.49} & \textbf{85.78} & \textbf{90.44} & \textbf{63.85} \\
    \bottomrule
    \end{tabular}%
      \caption{Perforamce comparison with the existing methods on RegDB dataset under visible-
infrared and infrared-visible settings. Rank at r accuracy (\%) and mAP (\%) are reported.}
  \label{tab:RegDB}%
\end{table*}%

\subsection{Comparisons with SOTA in Cross-Modality Person Re-ID}
In this section, we compare the proposed method with a number of cross-modality re-ID methods,
including One-stream~\cite{wu2017rgb}, Two-stream~\cite{wu2017rgb}, Zero-Pad~\cite{wu2017rgb}, TONE~\cite{ye2018hierarchical}, HCML~\cite{ye2018hierarchical}, cmGAN~\cite{dai2018cross}, BDTR~\cite{ye2019bi}, BDTR~\cite{ye2019bi}, HSME~\cite{hao2019hsme},
D2RL~\cite{wang2019learning}, MAC~\cite{ye2020cross},  MSR~\cite{feng2019learning},  AlignGAN~\cite{wang2019rgb},    Xmodal~\cite{li2020infrared}, and DDAG~\cite{eccv20ddag}.
Table~\ref{tab:SYSU-MM01} and~\ref{tab:RegDB} summary the experimental results on the SYSU-MM$01$ and the RegDB datasets, respectively. 

Table~\ref{tab:SYSU-MM01} reports the experimental results on SYSU-MM01. Our proposed method achieves better overall performance than all the other methods in terms of
all the evaluation metrics. Specifically, in all-search mode, our method can obtain 56.93\% for Rank-$1$ and 55.10\% for mAP in the all-search mode, which surpass the second best approach (\emph{i.e.}, DDAG~\cite{eccv20ddag}) by 2.18\% and 2.08\%, respectively.
Similar results are observed in indoor-search mode. Compared to the current SOTA method, we achieve 5.19 points and
3.54 points improvement on Rank-1 and mAP, respectively. The above impressive performance
suggests that our method can learn better modality-sharable features from patch-level to object-level.

We also evaluate our method against existing competing approaches on RegDB in Table~\ref{tab:RegDB}. As shown in Table~\ref{tab:RegDB}, Our method always
outperforms others by large margins in different query settings. For the Visible
to Infrared mode, it reaches 73.06\% on Rank-1 and 67.81\% on mAP, 3.72\% and 4.35\% higher
than the current SOTA (DDGA), respectively. For Infrared
to Visible, the improvements are 2.43\% on rank-1 and 2.05\% on mAP. This indicates that our DF$^{2}$AM is robust to different evaluation
modes.

\subsection{Ablation Study}
Extensive experiments are conducted under
four different settings in Table~\ref{tab:ablation1} to evaluate each component of our proposed method: (1) The effectiveness of DF$^{2}$ module, (2) The effectiveness of AM module, (3) The necessity of taking Baseline into our method. All the experiments are conducted on SYSU-MM$01$ dataset with two evaluation modes.

\begin{table}[!t]
\small
  \centering
  \setlength{\tabcolsep}{1.5mm}{
    \begin{tabular}{cccrccccc}
    \toprule
          & \multicolumn{1}{l}{Settings} &       &       & \multicolumn{2}{c}{All Search} &       & \multicolumn{2}{c}{Indoor  Search}  \\
\cmidrule{1-3}\cmidrule{5-6}\cmidrule{8-9}    Baseline & DF$^{2}$    & AM    &       & r=1   & mAP   &       & \multicolumn{1}{c}{r=1} & \multicolumn{1}{c}{mAP} \\
    \midrule
    $\surd$     &       &       &     &  52.88   & 51.14 &     & 56.16 & 64.27  \\
    $\surd$     & $\surd$    &       &   &  55.61    & 53.90  &      & 61.32 &  67.58 \\
    $\surd$     &       & $\surd$    &    & 52.88    & 52.09  &      & 59.38 &  66.82 \\
                & $\surd$     & $\surd$     &       &   55.80    &  54.04   &       &  65.80 &  70.15  \\
    $\surd$     & $\surd$     & $\surd$     &       & \textbf{56.93}   & \textbf{55.10}  &    &   \textbf{66.39} & \textbf{71.52} \\
    \bottomrule
    \end{tabular}}%
      \caption{Ablation study. We evaluate five settings on SYSU-MM01 dataset. ``Baseline'', ``Baseline'' with DF$^{2}$, ``Baseline'' with AM, DF$^{2}$ with AM, and our DF$^{2}$AM.  Our method achieves the best result among other competitors.}
  \label{tab:ablation1}%
\end{table}%

As shown in Table~\ref{tab:ablation1}, single-modality re-ID
 model obtain a better result than current some method. This indicates that some training
tricks taken from single-modality Re-ID [67] also contributes to the performance of our method.
Next, we directly merge DF$^{2}$ into the training process based on baseline model,
which improves mAP scores by 2.76\% and 3.31\% for both evaluation modes. The above impressive improvements demonstrate that learning
dual-level features is beneficial for cross-modality Re-ID. Similar results can be observed when AM is integrated into the training process,
which also demonstrates the effectiveness of AM module.
As the fifth line of Table~\ref{tab:ablation1} shows,
the performance is further improved when we aggregate two modules with
baseline. The impressive improvement
suggests that these components are mutually beneficial to each other. However, without baseline, the performance of our method has a slight decline, demonstrating the necessity of taking Baseline into our method.

\begin{table}[t]
\small
  \centering
    \begin{tabular}{c|c|c|c|c}
    \toprule
    Method & mAP   & r=1   & r=5   & r=10 \\
    \midrule
   DF$^{2}$AM(3 parts) & 68.06 & 62.45 & 91.71 & 96.83 \\
    DF$^{2}$AM(4 parts) & \textbf{71.52} & \textbf{66.39} & \textbf{94.93} & \textbf{98.55} \\
    DF$^{2}$AM(5 parts) & 70.40  & 65.17 & 92.93 & 97.69 \\
    \bottomrule
    \end{tabular}%
          \caption{The performance of proposed DF$^{2}$AM with different number
of splitted feature parts when trained on SYSU-MM$01$ dataset and tested
on indoor search mode.}
  \label{tab:parts}%
\end{table}%

In addition, we compare the performance of DF$^{2}$AM
with different number of sliced horizontal feature
parts. As shown in Table~\ref{tab:parts}, our method achieves the best results
when the feature maps are splitted into four parts.
In this way, local features contain the most discriminative information for cross-modality re-ID.

\begin{figure}
\begin{center}
\includegraphics[width=1\columnwidth]{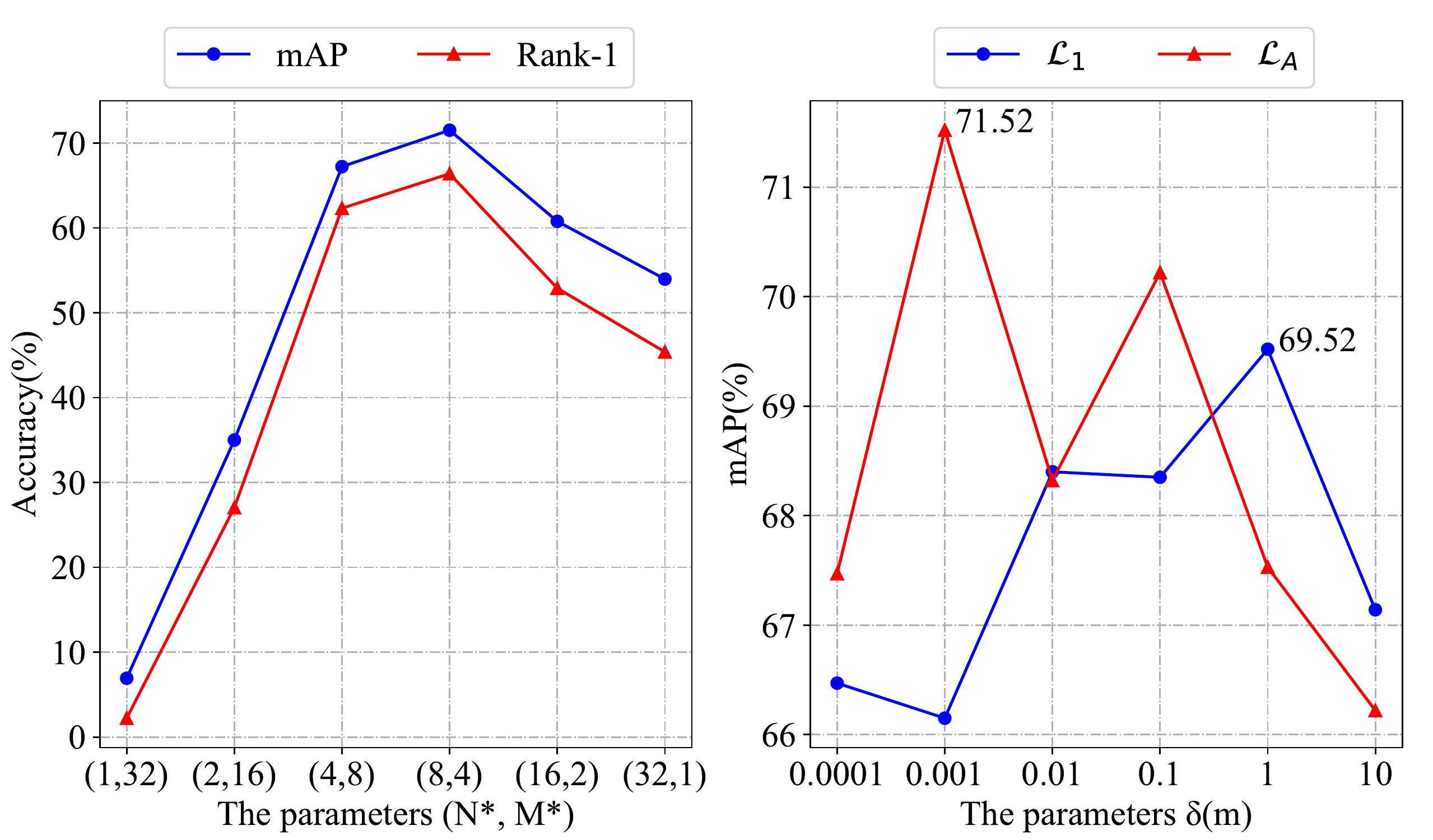}
\end{center}
   \caption{\small The experimental results on SYSU-MM$01$ under indoor-search mode. Left: The performance along with different parameters $(N,M)$. Right:
   The performance of our method with different loss functions under varying values of parameters $\delta(m)$.
}
\label{fig:nm}
\end{figure}

\textbf{How do the identity and sample numbers of sampling strategy affect representation quality?} The hyper-parameter $N^{*}$ and $M^{*}$ controls identity number and sample numbers for each identity in training batch, respectively.
For fair comparison, batch size $2K^{*}$ is fixed as $64$, while $K^{*}$, $N^{*}$ and $M^{*}$ satisfy the condition $K^{*}=M^{*}N^{*}$.
Figure~\ref{fig:nm} shows how the re-ID performance varies with different
numbers of person identities, $N^{*}$. It can be observed that as $N^{*}$
increases from 1 to 8, the re-ID performance continues rising,
and the performance begins to decrease when $N^{*}$ gets larger. That is, the best accuracy is achieved with $N^{*}=8$ and $M^{*}=4$. The main reason is when $N^{*}$ is too small, the learned similarity
is not adequate, which makes the model
difficult to match features of the same identity during affinity modeling. When $N^{*}$ is
too large, images of different persons will be reduced, which harms the network training for similarity inference.

\textbf{The impact of $\mathcal{L}_A$ and $\mathcal{L}_1$}. As shown in Figure~\ref{fig:nm}, our method (w/ $\mathcal{L}_A$) achieves the best performance of mAP=71.52\% on SYSU-MM$01$,
2.00\% higher than directly using $\mathcal{L}_1$. Such results prove the necessity and effectiveness
of our affinity modeling loss $\mathcal{L}_A$, which alleviates the problem of model converging to local minima early in training stage. In addition, it can be observed that compared with our method (w/ $\mathcal{L}_1$), smaller value $m(\delta)$ leads to the better results of our method (w/ $\mathcal{L}_A$) and vice versa, which is consistent with our expectation. The main reason
is that our network is trained with $\mathcal{L}_A$, which
avoids pushing always images of different
identities separate as $\mathcal{L}_1$, but it has a relatively small strength (\emph{i.e.}, $m$) to force them separate.

\begin{figure}
\begin{center}
\includegraphics[width=1\columnwidth]{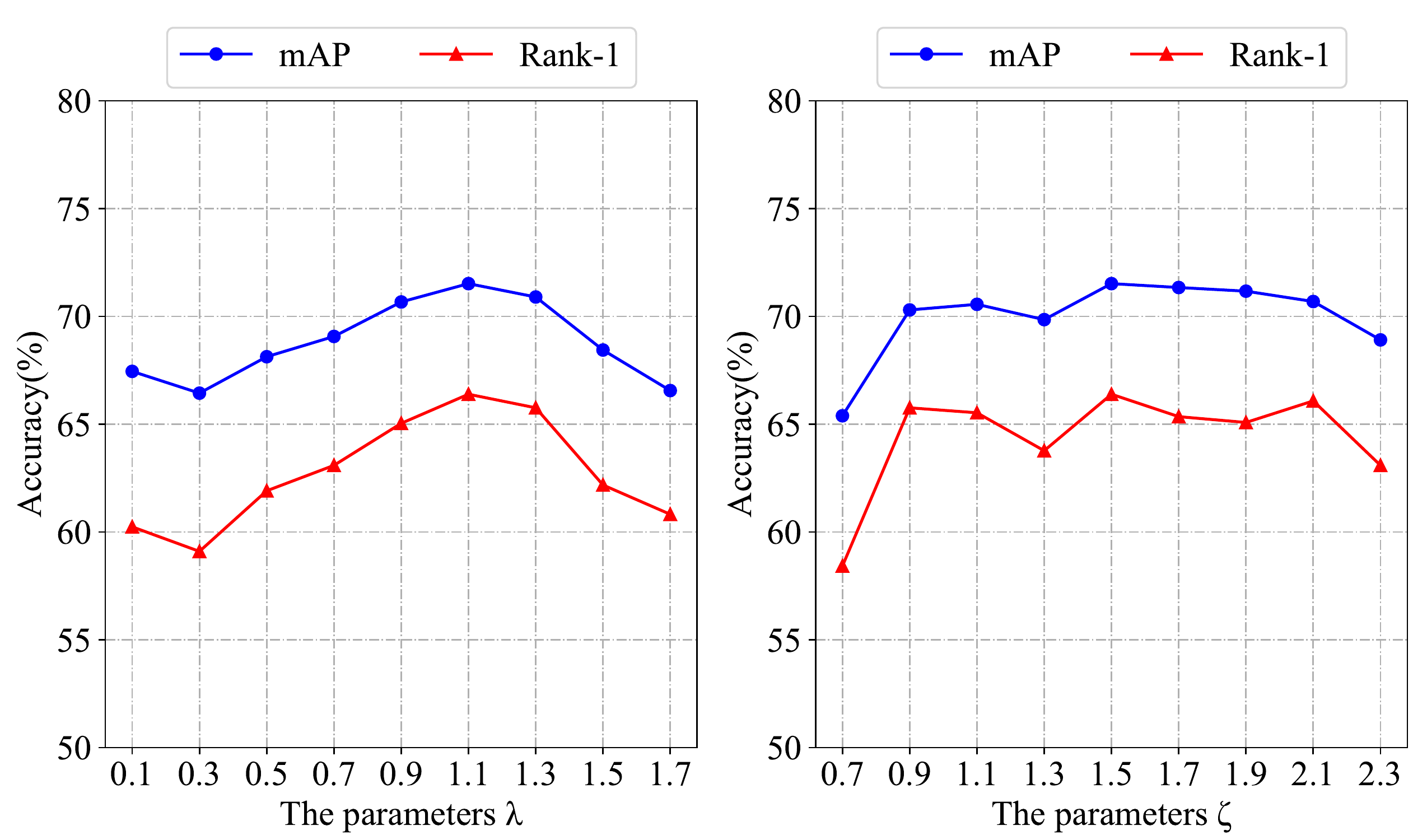}
\end{center}
\caption{\small Performance evaluation of our proposed method with different values of $\lambda$ and $\zeta$
 on SYSU-MM$01$ under indoor-search mode. 
}
\label{fig:para}
\end{figure}

\subsection{Parameter sensitivities}
The parameter $\lambda$ balances the effect of DF$^{2}$ module
and baseline model. We evaluate our method with different values for
the parameter $\lambda$ in Figure~\ref{fig:para}. As $\lambda$ increases, the accuracy
improves at first. When $\lambda=1.1$, we obtain the best
performance. After that, the performance begins to decline. Similar results can be observed when
we vary parameter values of $\zeta$ from 0.1 to 2.1.
The optimal accuracy is achieved when $\zeta=1.5$. Despite the performances vary with different parameter values, most results by our approach
outperform the current state-of-the-arts significantly.

\section{Conclusions}
In this paper, we propose an efficient network that integrates
feature information at different modalities into person re-identification.
we introduce two key modules into the network and fuse dual-level feature to reduce the cross-modality discrepancy.
The proposed DF$^{2}$ module
considers both the
global feature and its part information to determine importance between different body parts of a person from global
and local views. The AM module uses intra-class compactness and inter-class
separability in the sample similarities as supervised information to model the relationships between person images.
Ablation studies demonstrate the effectiveness
of the proposed modules in improving the identification accuracy.
Extensive experiments with our
state-of-the-art results on two competitive datasets further demonstrate
the effectiveness and generality of our DF$^{2}$AM approach.

{\small
\bibliographystyle{ieee_fullname}
\bibliography{egbib}
}

\end{document}